\documentclass{article}

\usepackage{PRIMEarxiv}

\usepackage[utf8]{inputenc} 
\usepackage[T1]{fontenc}    
\usepackage{hyperref}       
\usepackage{url}            
\usepackage{booktabs}       
\usepackage{amsfonts}       
\usepackage{nicefrac}       
\usepackage{microtype}      
\usepackage{lipsum}
\usepackage{graphicx}
\graphicspath{{media/}}     

 \usepackage{amsmath}
\usepackage{balance}
\usepackage[dvipsnames]{xcolor}
\usepackage{multirow} 

\newcommand{\datasetname}{\textit{``MisCaption This!''}}
  
\title{Latent Reconstruction from Generated Data for Multimodal Misinformation Detection}

\usepackage{authblk}

\author[1, 2]{\small Stefanos-Iordanis Papadopoulos\thanks{Corresponding author} \ }
\author[1]{\small Christos Koutlis}
\author[1]{\small Symeon Papadopoulos}
\author[2]{\small Panagiotis C. Petrantonakis}

\affil[1]{\footnotesize Information Technology Institute, Centre for Research \& Technology, Hellas.}
\affil[2]{\footnotesize Department of Electrical \& Computer Engineering, Aristotle University of Thessaloniki.}
\affil[ ]{\textit {\{stefpapad,ckoutlis,papadop\}@iti.gr, \textit{ppetrant@ece.auth.gr}}}

\begin{document}
\maketitle

\begin{abstract}
Multimodal misinformation, such as miscaptioned images, where captions misrepresent an image’s origin, context, or meaning, poses a growing challenge in the digital age.
Due to the scarcity of large-scale annotated datasets for multimodal misinformation detection (MMD), recent approaches rely on synthetic training data created via out-of-context pairings or named entity manipulations (e.g., altering names, dates, or locations).
However, these often yield simplistic, unrealistic examples, which limits their utility as training examples. 
To address this, we introduce \datasetname, a framework for generating high-fidelity synthetic miscaptioned datasets through Adversarial Prompting of Vision-Language Models (VLMs).
Additionally, we introduce ``Latent Multimodal Reconstruction'' (LAMAR), a Transformer-based network trained to reconstruct the embeddings of truthful captions, providing a strong auxiliary signal to guide detection. 
We explore various training strategies (end-to-end vs. large-scale pre-training) and integration mechanisms (direct, mask, gate, and attention). 
Extensive experiments show that models trained on \datasetname\ data generalize better to real-world misinformation, while LAMAR achieves new state-of-the-art on NewsCLIPpings, VERITE, and the newly introduced 
VERITE 24/25 benchmark; highlighting the efficacy of VLM-generated data and reconstruction-based networks for advancing MMD.
Our code is available at \url{https://github.com/stevejpapad/miscaptioned-image-reconstruction}.
\end{abstract}

\keywords{Multimodal Learning \and Deep Learning \and Misinformation Detection \and Reconstruction Network}

\section{Introduction}
\label{sec:intro}

The rise of the Internet and digital technologies has significantly accelerated the spread of  information but also the proliferation of misinformation, including new forms of deceptive content such as DeepFakes \cite{rana2022deepfake}, multimodal misinformation \cite{akhtar2023multimodal}, and LLM-generated misinformation \cite{chen2024combating}.
Given the scale and speed of misinformation dissemination, researchers are developing automated fact-checking tools to assist human fact-checkers in identifying deceptive content more efficiently \cite{guo2022survey}.
In this study, we focus on multimodal misinformation detection (MMD), specifically, the detection of misleading image-caption pairs, where both modalities jointly contribute to the spread of false or misleading information \cite{alam2022survey}.

Due to the scarcity of large-scale, expert-annotated MMD datasets, researchers have increasingly turned to synthetic training data. 
While fact-checking archives provide high-fidelity labels, their limited volume of multimodal pairs, lack of negative samples, and licensing constraints render them useful for evaluation but insufficient for training. 
Consequently, aside from a few small-scale annotated resources \cite{zlatkova2019fact}, recent studies have had to rely on either weakly annotated \cite{boididou2018verifying, nakamura2020fakeddit, nielsen2022mumin, jindal2020newsbag} or algorithmically generated training datasets; created either by pairing images with out-of-context (OOC) captions from other images \cite{jaiswal2017multimedia, aneja2023cosmos, biamby2022twitter, luo2021newsclippings} or by manipulating named entities to introduce inconsistencies, resulting in miscaptioned (MC) images \cite{sabir2018deep, muller2020multimodal, papadopoulos2023synthetic}.
However, named-entity manipulation often produces shallow or implausible misinformation, lacking the nuance seen in real-world examples. 
At the same time, the potential of Vision Language Models (VLMs) to generate richer, more diverse synthetic training data remains underexplored.

On the modeling side, recent works have explored large pre-trained encoders \cite{luo2021newsclippings}, self-supervised fine-tuning \cite{mu2023self}, attention-based fusion mechanisms \cite{kumari2021amfb, yu2022bcmf, papadopoulos2023red}, external evidence \cite{abdelnabi2022open, yuan2023support}, and Vision-Language Models (VLMs) for both detection \cite{qi2024sniffer, tahmasebi2024multimodal} and explanation generation \cite{zhang2023ecenet}.
However, while reconstruction networks have shown promise in other domains, their use in MMD is still limited -- despite their alignment with human fact-checking strategies, where reconstructing the original meaning, context, or source of an image is key to detecting misinformation \cite{cazzamatta2025decoding}.

\begin{figure*}[t]
\centering
\includegraphics[width=\textwidth]{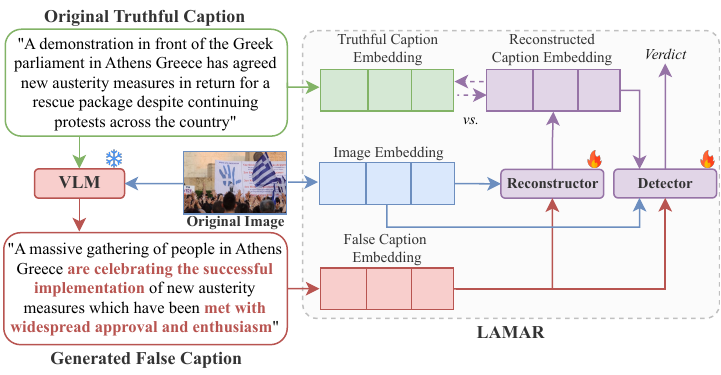}
\caption{High-level overview of the proposed workflow. 
A VLM generates a false caption from the original, truthful image-caption pair. 
The Reconstructor (Transformer encoder) then takes both the image and false caption embeddings as inputs to identify and rectify inaccuracies in the text, recreating the original truthful caption embedding. 
This reconstructed representation is fused with the other modalities via a specialized mechanism (e.g., Gating or Attention) and passed to the detector to produce the final verdict. 
The network is trained to simultaneously minimize the error between original and reconstructed embeddings and optimized for classification accuracy.}
\label{fig:high_level}
\end{figure*}

To this end, we propose \datasetname, a framework that leverages VLMs to manipulate the captions of images drawn from truthful image-caption datasets, generating \textit{synthetic} false captions that misrepresent aspects of the image.
The resulting datasets are used to train the proposed Latent Multimodal Reconstruction (LAMAR) network, which learns to reconstruct the embedding of the original, truthful caption.
As shown in Fig.~\ref{fig:high_level}, LAMAR takes the image and manipulated caption embeddings as input and is tasked with recovering the original caption embedding.
This reconstructed embedding is then integrated into the final detection network alongside the fused multimodal representation.

Our rationale is that (1) VLMs can generate more diverse and realistic synthetic training data, and (2) semantic divergence between reconstructed and input embeddings provides a discriminative signal that exposes cross-modal and intra-modal inconsistencies, significantly improving model generalization.

\datasetname\ uses open-source VLMs (LLaVA, Llama, Molmo) and apply adversarial prompt selection, where generative prompts are evaluated against the VLM’s zero-shot detection capabilities. Prompts that produce easily detectable or overly generic captions are discarded. 
LAMAR employs a Transformer encoder with element-wise modality fusion, and we explore both end-to-end training and large-scale pre-training. We also compare four integration strategies for the reconstructed embedding: direct integration, masking, gating, and self-attention.

Our main contributions can be summarized as follows:
\begin{itemize}

\item We introduce \datasetname, a framework for creating synthetic training datasets of miscaptioned images using open-source VLMs.

\item We propose the Latent Multimodal Reconstruction (LAMAR) architecture, which learns to reconstruct the truthful caption embedding from its input.

\item 
We show that models trained on VLM-generated data generalize better to real-world misinformation, outperforming named-entity swapping and cross-modal misalignment datasets by 7.8\% and 10.4\% on the VERITE benchmark \cite{papadopoulos2024verite}, respectively. 

\item LAMAR achieves new state-of-the-art (SotA) results on NewsCLIPpings and across all VERITE tasks, with improvements of 4.3\% on ``True vs. MC'', 3.0\% on ``True vs. OOC'', and 5.6\% on multiclass classification.

\item LAMAR demonstrates stronger temporal generalization and robustness on emerging misinformation compared to zero-shot VLMs, 
maintaining high performance (70.0\%) on the new $\textit{VERITE 24/25}$ benchmark -- consisting of events after the VLM knowledge-cutoff -- while VLM performance degrades significantly (8--12\% reduction).

\end{itemize}

\section{Related Work}
\label{sec:rw}

\subsection{MMD Datasets}

Training machine learning models for MMD requires suitable datasets, with current research focusing on annotated, weakly annotated, and synthetically generated data.
Early MMD datasets, such as the \textit{Twitter}  \cite{boididou2018verifying} and \textit{Weibo} \cite{jin2016novel} datasets, are relatively small and cover only a limited number of events-17 and 73, respectively-raising concerns about model generalization. 
To address this, larger weakly annotated datasets have emerged, including \textit{MuMiN} \cite{nielsen2022mumin}, with rich social context but few images, and \textit{NewsBag} \cite{jindal2020newsbag}, which includes satirical content, and \textit{Fakeddit} \cite{nakamura2020fakeddit}, with over a million instances collected from Reddit.
However, studies show that models trained on \textit{Twitter} and \textit{Fakeddit} often exhibit unimodal biases, undermining their effectiveness in real-world multimodal misinformation detection \cite{papadopoulos2024verite}.

Researchers have also explored synthetic data generation. 
These approaches can be categorized as either out-of-context (OOC) pairs or named entity swapping (NES). 
Early OOC datasets, like \textit{MAIM} \cite{jaiswal2017multimedia} and \textit{COSMOS} \cite{aneja2023cosmos}, used random image-text mismatches, which often resulted in unrealistic and easy to detect samples \cite{papadopoulos2023synthetic}. 
More refined approaches, such as \textit{NewsCLIPings} \cite{luo2021newsclippings} and \textit{Twitter-COMMs} \cite{biamby2022twitter}, incorporated CLIP-based retrieval to enhance cross-modal relevance. 

NES-based datasets generate misinformation by replacing named entities in captions with alternatives retrieved from similar or contextually relevant texts using 
cluster-based retrieval (\textit{MEIR} \cite{sabir2018deep}), rule-based substitutions (\textit{TamperedNews} \cite{muller2020multimodal}), 
and CLIP-based retrieval (\textit{CLIP-NESt} \cite{papadopoulos2023synthetic}).
Despite recent progress in VLMs and their use for synthetic dataset generation \cite{li2023synthetic}, their application to MMD remains limited. 
To date, only \textit{MMFakeBench} \cite{liu2024mmfakebench} uses VLM-generated rumors and AI-manipulated images, and serves solely as a small-scale evaluation benchmark. 
In contrast, we introduce a large-scale VLM-generated training dataset for MMD.

However, models trained and evaluated on synthetic data may struggle with real-world generalizability, as they may learn to detect patterns specific to artificially generated inconsistencies rather than the more complex and diverse manipulations found in real-world misinformation.
To address this, benchmarks like \textit{VERITE} incorporate real-world OOC and miscaptioned images \cite{papadopoulos2024verite}.

\subsection{MMD Methods}

Research on MMD has centered on developing models that encode textual and visual modalities, fuse their representations, and assess their consistency and factual accuracy.
Early methods, such as SpotFake \cite{singhal2019spotfake}, used VGG-19 and BERT, 
while recent approaches use pre-trained modality-specific dual-encoder models such as CLIP \cite{luo2021newsclippings}, or fine-tune it through self-supervised learning \cite{mu2023self}. 
Some models incorporate multi-task learning, such as EANN with an event discriminator \cite{wang2018eann} or MVAE, which uses an autoencoder to reconstruct the input text and visual features \cite{khattar2019mvae}; but does not modify the input text or reconstruct truthful captions from false ones. 

While earlier methods relied on simple concatenation of visual and textual embeddings, more advanced approaches have explored Attention-based Multimodal Bilinear Pooling \cite{kumari2021amfb}, Bidirectional Crossmodal Fusion (BCMF) \cite{yu2022bcmf}, multi-head attention in Transformers \cite{papadopoulos2023synthetic}, and element-wise vector fusion \cite{papadopoulos2023red} to enhance cross-modal interaction.
Recent work integrates external web evidence, with methods assessing internal and external consistency (CCN \cite{abdelnabi2022open}, SNIFFER \cite{qi2024sniffer}) or evaluating stance and relevance of external evidence (SEN \cite{yuan2023support}, RED-DOT \cite{papadopoulos2023red}). 
While external evidence is shown to improve performance, concerns remain about `leaked evidence' from fact-checking articles \cite{glockner2022missing, chrysidis2024credible} and datasets-specific artifacts that models may exploit instead of assessing factuality \cite{papadopoulos2024similarity}. For these reasons, we do not consider evidence-based approaches in this study. 

\subsection{Reconstruction Networks}

Reconstruction networks are deep learning models designed to generate or restore original data from low-resolution or altered inputs. 
They have been applied in various domains, 
including few-shot image classification by reframing it as a reconstruction problem in latent space \cite{wertheimer2021few}, 
image inpainting through an adversarial framework guided by textual descriptions \cite{wu2021adversarial}, 
super-resolution reconstruction to enhance low-resolution images of the same scene \cite{yan2024research}, 
and DeepFake detection \cite{he2022defeating}.
Reconstruction networks remain largely unexplored in MMD, with MVAE \cite{khattar2019mvae} being a notable exception. 
However, MVAE employs a variational autoencoder to reconstruct the input, recovering the input words (via Bi-LSTMs) and VGG-19 image embeddings from a shared latent space. 
Crucially, this reconstruction serves only to enhance feature extraction and does not involve modifying or correcting the input content, e.g., generating truthful descriptions, or their corresponding latent representations, from false ones. 

\section{Problem Formulation}

Given a set $\mathcal{D}^t = {(I^t_i, C^t_i)}_{i=1}^{N}$ of $N$ image-caption pairs, where each $I^t_i$ is an image and $C^t_i$ its matching, truthful caption, we define:

\noindent 
\textbf{Out-of-Context (OOC)} pairs as $(I^t_i, C^x_i)$, where $C^x_i$ is a caption taken from a different sample in $\mathcal{D}^t$;

\noindent 
\textbf{Mis-Captioned (MC)} pairs as $(I^t_i, C^f_i)$, where $C^f_i$ is a manipulated version of the original caption that misrepresents the content or meaning of $I^t_i$.

A ``manipulator'' refers to any method used to create OOC (e.g., CLIP-based retrieval) or MC (e.g., entity swapping) samples, producing datasets $\mathcal{D}^x$ and $\mathcal{D}^f$ from $\mathcal{D}^t$.

We define MMD as a classification task to learn a mapping $\mathsf{M}^d: \mathcal{D} \rightarrow \hat{y}$, where $\hat{y}$ is the predicted class label under one of three settings:

\noindent 
(1) \textbf{Binary: `True vs. MC'}, where $\mathcal{D} = [\mathcal{D}^t, \mathcal{D}^f]$ with $K = N * 2 $ total pairs and $y \in \{ 0, 1 \}$ 

\noindent 
(2) \textbf{Binary: `True vs. OOC'}, where $\mathcal{D} = [\mathcal{D}^t, \mathcal{D}^x]$ with $K = N * 2 $ total pairs and $y \in \{ 0, 2 \}$, 

\noindent 
(3) \textbf{Multi-class: `True vs. MC vs. OOC'}, where $\mathcal{D} = [\mathcal{D}^t, \mathcal{D}^x, \mathcal{D}^f]$ with $K = N * 3$ total pairs and $y \in \{ 0, 1, 2\}$.

Let $\mathsf{E}_I$ and $\mathsf{E}_C$ denote the image and text encoders, producing embeddings $\mathbf{I}$ and $\mathbf{C}$ for input pair $(I, C)$, respectively.
We define latent reconstruction as learning a function $\mathsf{M}^r: (\mathbf{I}, \mathbf{C}) \rightarrow \hat{\mathbf{C}}^t$, where $\hat{\mathbf{C}}^t$ approximates the embedding of the original truthful caption $\mathbf{C}^t$.
The model is trained to minimize reconstruction loss:
\begin{equation}
\label{eqn:loss}
\mathcal{L}_r(C^t, \hat{C}^t) = \frac{1}{K} \sum_{i=1}^{K} (\mathbf{C}^t_i - \hat{\mathbf{C}}^t_i)^2
\end{equation}
where $\mathcal{L}_r$ is the Mean Squared Error (MSE) between the true and reconstructed caption embeddings.

Unless explicitly denoted as $C^t$ (truthful), $C^f$ (falsified), or $\mathbf{\hat{C}}^t$ (reconstructed), we use $C$ and $\mathbf{C}$ to refer to the input caption and its embedding--regardless of truthfulness.

\section{\datasetname}
\label{sec:dataset}

In this study, we explore the creation of synthetic training datasets of miscaptioned images ($\mathcal{D}^f$) by manipulating the image captions of a truthful dataset ($\mathcal{D}^t$) using a VLM as the ``Manipulator''.
Our rationale is that VLMs, with their advanced multimodal understanding and generation capabilities, can produce more realistic false captions for images compared to methods relying on manipulating named entities. 
In turn, we hypothesize that the generated data $\mathcal{D}^f$ can be leveraged to train more robust detection models $\mathsf{M}^d$, thus enhancing generalization to real-world misinformation.

\subsection{Generative Model}
To generate synthetic training data, we employ open-source VLMs:
LLaVA-1.6-Mistral\footnote{\url{https://huggingface.co/LLaVA-hf/LLaVA-v1.6-mistral-7b-hf}}, Molmo-7B-D\footnote{\url{https://huggingface.co/allenai/Molmo-7B-D-0924}}, and Llama-3.2-11B-Vision\footnote{\url{https://huggingface.co/meta-llama/Llama-3.2-11B-Vision}}.
We also explored LLaVA-1.6-Vicuna-13B, Janus Pro 7B (DeepSeek), and MiniGPT-v2-Llama-2-7B as alternatives, however, we were unable to get them to consistently generate realistic false captions, as they often defaulted to generic image captioning, re-phrasing of the original caption, or overly simplistic misinformation. 
In contrast, GPT-4o Mini demonstrated robust safeguards, often refraining from generating misinformation altogether.

\subsection{Adversarial Prompt Selection}

Generating realistic misinformation with an VLM requires carefully selected generative prompts ($p^{gen}$). 
However, identifying an ``optimal'' prompt for a task is non-trivial and costly to evaluate exhaustively. 
To mitigate this, we leverage \textit{Adversarial Prompt Selection}, which filters prompts based on how well their generated captions ($C^f$) evade detection by a zero-shot VLM, ensuring the generated data presents a meaningful challenge.

First, we assess the VLM's (LLaVA) zero-shot capability as a `Detector' of miscaptioned images.
We conduct a grid search over a candidate pool of prompts varying in wording and granularity. 
These were evaluated on a balanced calibration set ($N=100$) from the VERITE (``True vs. MC'') benchmark.
Through this process, we identified the optimal prompt as: ``The image is captioned as: [CAPTION]. Is the caption truthful or does it contain falsehoods?'', which achieved $\approx70\%$ accuracy. 
This prompt was selected for subsequently experiments as the adversary detection prompt ($p^{dt}$) against which we evaluate candidate generative prompts ($p^{gen}$).
Our analysis revealed that the specific phrasing ``contain falsehoods'' consistently outperformed synonyms such as ``manipulated'', ``fake'', or ``falsified''. 
Furthermore, explicitly defining terms, such as stating that false captions ``may falsely describe origin, context, and/or meaning of the image'' or ``contain inconsistencies between named entities (people, dates, locations etc.)'' hindered performance, yielding only $\approx62\%$ accuracy.

\begin{figure*}[!t]
\centering
\includegraphics[width=\textwidth]{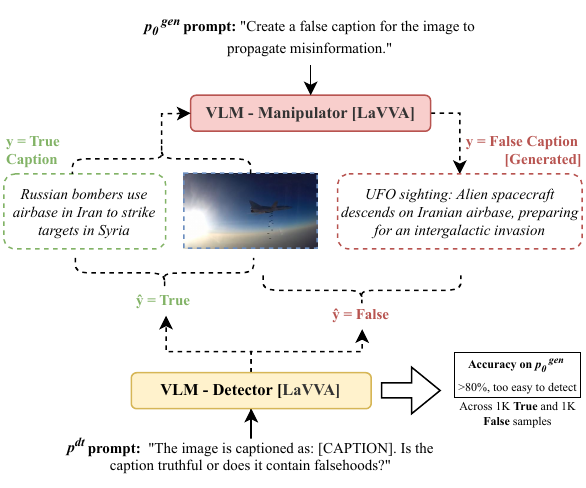}
\caption{Adversarial Prompt Selection: An VLM `Manipulator' generates a false caption using the generative prompt $p_0^{gen}$. 
The VLM `Detector' is then evaluated on the zero-shot classification of both the truthful and generated captions.
Intermittent lines indicate the prediction on a single sample, while the overall accuracy is calculated across a balanced set of 2,000 samples. 
In this specific case, the high overall detection accuracy indicates the generated misinformation is too simplistic; thus $p_0^{gen}$ is not selected for the creation of a training dataset.
}
\label{fig:adv_prompting}
\end{figure*}

Next, we assess a range of generative prompts by using the VLM as `Manipulator' to produce 1,000 falsified captions ($C^f_j$) for 1,000 randomly sampled truthful examples from NewsCLIPpings, as expressed in Eq. \ref{eqn:gen}. 
Generated captions ($C^f_j$) are assigned the ground truth label `False'. 
To evaluate the quality of the generated misinformation, we then subject both the generated captions ($C^f_j$) and the original, truthful captions ($C^t$ - labeled `True') to zero-shot classification by the VLM-Detector using the selected detection prompt, as expressed in Eq. \ref{eqn:zero_shot}.

\begin{equation}
\label{eqn:gen}
C^f_{j} = \text{VLM-Manipulator}(I, C^t \mid p^{gen}_j), \quad \forall j \in [1, J]
\end{equation}
\begin{equation}
\label{eqn:zero_shot}
\hat{y}_{j} = \text{VLM-Detector}(I, C_j \mid p^{dt}), \quad C_j \in \{C^t_j, C^f_j \}
\end{equation}

Prompts that result in high detection accuracy often generate simplistic or unrealistic misinformation. 
For example, as seen in  Fig. \ref{fig:adv_prompting}, the prompt ``Create a false caption for the image to propagate misinformation'' produced: ``UFO sighting: Alien spacecraft descends on Iranian airbase, preparing for an intergalactic invasion'' for this specific image, and the VLM-Detector yielded high detection accuracy (83.5\%).
Conversely, prompts yielding very low detection accuracy could mean that the generated captions are truthful, generic or re-phrased captions, failing to introduce actual falsehoods.
For example, the generic prompt ``Provide a caption for the image'' produced ``A lone fighter jet soaring through the clear blue sky, leaving a trail of contrails behind'' for the same image, with low overall detection accuracy (42.5\%).  
Therefore, we do not select prompts that produce either easily detectable false captions (due to being unrealistic) or those that fail to introduce falsehoods (due to either being truthful or generic descriptions).

Through this process and manual inspection, we select four generative prompts ($p^{gen}_1$ – $p^{gen}_4$) that span a range of adversarial difficulty (57\%, 62\%, 71\%, 75\% detection accuracy). 
These are used to generate four datasets ($\mathcal{D}^f_1$–$\mathcal{D}^f_4$) of falsified captions conditioned on both the image and its original caption. 
After identifying the most effective generative prompt ($p^{gen}_3$) for LLaVA, we use it for Molmo and Llama as well to keep the computational load of our experiments to reasonable levels while testing cross-model generalizability.
For ethical reasons, we withhold exact prompts but make them available upon request under a research-only license.

\begin{figure*}[!t]
\centering
\includegraphics[width=\textwidth]{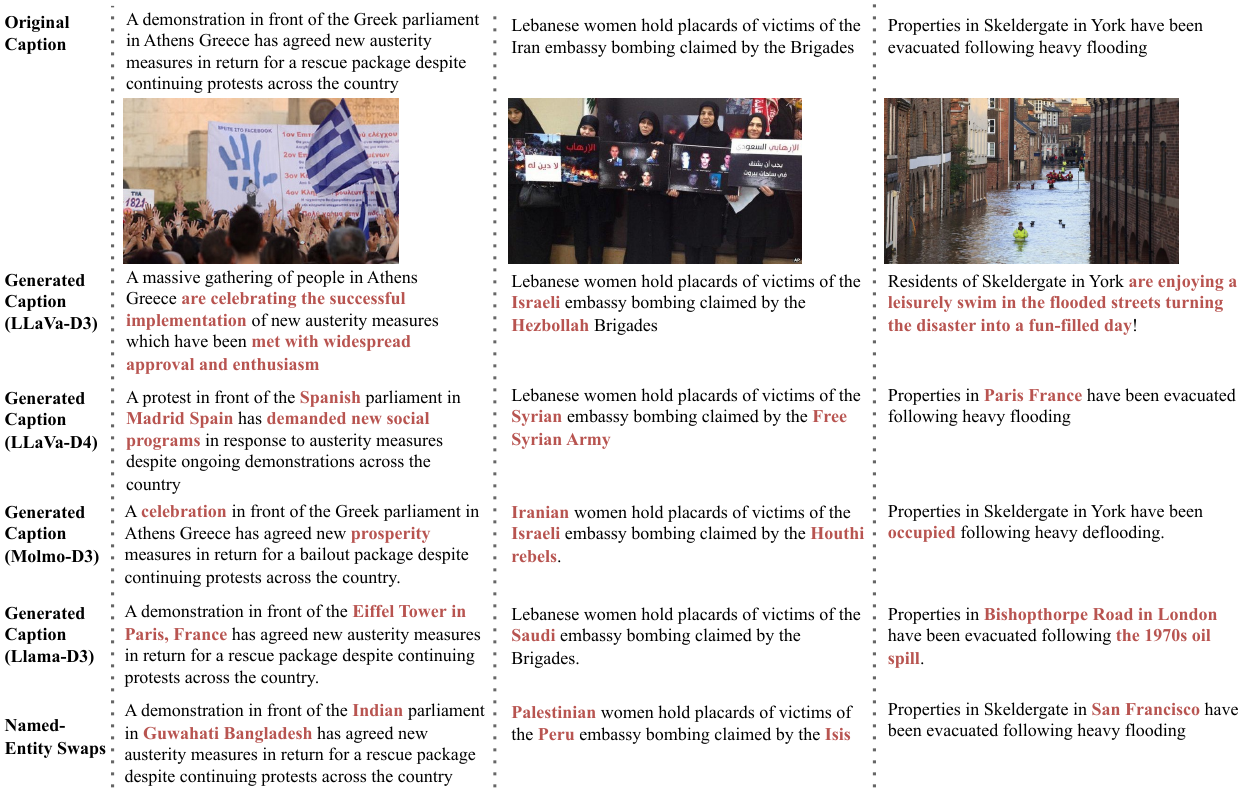}
\caption{Examples of truthful and generated captions, alongside false captions created via named entity swaps.}
\label{fig:examples}
\end{figure*}

\subsection{Generated Data Filtering}
Qualitative analysis (such as in Fig.~\ref{fig:examples}) showed that VLMs can produce more coherent and plausible false captions than NES-based methods, especially in terms of logical and factual consistency.
For instance, NES manipulations yielded implausible claims like the `Indian parliament' being in `Bangladesh' or `Skeldergate' in `San Francisco', highlighting the limits of surface-level entity substitution.
In contrast, VLMs showed the ability to generate more nuanced and believable misinformation. 
It might, for instance, reframe an anti-austerity protest as pro-austerity, reinforcing the false narrative with phrases like ``successful implementation'' and ``widespread approval'', or distort the context of flood evacuations by suggesting that people were ``enjoying a leisurely swim'' and ``turning a disaster into a fun-filled day''.

Nevertheless, LLaVA occasionally ``rambles'', rephrasing the original caption with superficial elaborations rather than introducing actual misinformation.
For instance, from the original caption:
``The recent economic boom has enabled new projects such as the Union Trade Centre shopping centre in the heart of Kigali'', LLaVA generated:
``The Union Trade Centre shopping centre in Kigali is a prime example of the city's thriving economy with numerous cars parked outside indicating a bustling shopping scene'', 
which merely rephrases the original with extraneous details.
Moreover, if generation often results in longer captions, this could introduce superficial patterns (``shortcuts''), artifacts, that detection models may learn to exploit, instead of learning deeper features that are associated with content factuality and credibility.

To address this, we apply a post-processing filter that excludes samples where the generated caption exceeds the original by a relative character-length threshold $l \in \{0, 5, 10, 15, 25, 50, \text{None}\}$, retaining 4.5\%, 19.1\%, 27.8\%, 34.9\%, 47.9\%, 74.0\%, and 100\% of the LLaVA-$\mathcal{D}_3$ dataset, respectively. 
We empirically evaluate how filtering affects model performance.
To maintain class balance, we removed both the generated pair $(I^t_i, C^f_i)$ and its corresponding truthful pair $(I^t_i, C^t_i)$ when this threshold is exceeded.

\subsection{Dataset Source and Statistics}
We use the NewsCLIPpings `Merged/Balanced' dataset \cite{luo2021newsclippings}, 
which has been shown to be effective for OOC detection \cite{abdelnabi2022open, qi2024sniffer, papadopoulos2023red, papadopoulos2024similarity}, 
as the source dataset $\mathcal{D}^t$ from which to generate $\mathcal{D}^f$, utilizing only its truthful data: 35,536 pairs for training, 3,512 for validation, and 3,512 for testing.
After generating $\mathcal{D}^f_1$, $\mathcal{D}^f_2$, $\mathcal{D}^f_3$, and $\mathcal{D}^f_4$ from $\mathcal{D}^t$ using LLaVA and the corresponding generative prompts, we merge $\mathcal{D} = [\mathcal{D}^t, \mathcal{D}^f]$ to address the ``True vs. MC'' task, 
or integrate the full NewsCLIPpings dataset $\mathcal{D} = [\mathcal{D}^t, \mathcal{D}^x, \mathcal{D}^f]$ to address the multi-class task.
We preserve the original train/validation/test split of the NewsCLIPpings dataset to prevent any data leakage.
This results in a total of 106,605, 10,536, and 10,896 samples for training, validation, and testing, respectively, ensuring a balanced distribution across the three classes.
We use the Molmo and Llama generated datasets in the same manner.

\section{Latent Multimodal Reconstruction (LAMAR)}

\begin{figure}[!t]
\centering
\includegraphics[width=0.45\textwidth]{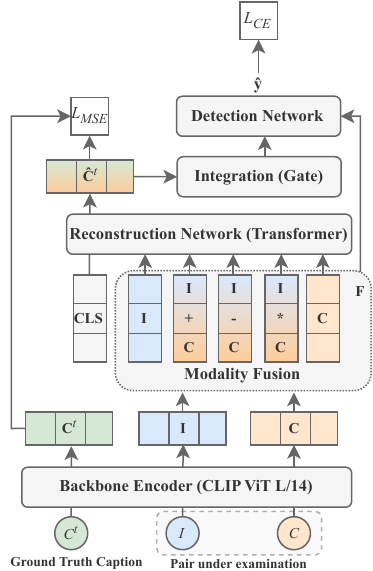}
\caption{
End-to-end training of the proposed LAMAR architecture.
The input caption $C$ may be either truthful or falsified. A CLIP ViT-L/14 encoder and a Transformer-based reconstruction module, enhanced with element-wise vector operations for modality fusion, outputs the reconstructed embedding $\mathbf{\hat{C}^t}$. 
This is integrated into the detection network via some mechanisms (e.g., gate, mask, attention), which outputs the final verdict. 
The reconstruction module is trained with MSE loss against the ground-truth $\mathbf{C}^t$, while the detection module is trained with cross-entropy (CE).
}
\label{fig:lamar}
\end{figure}

We propose a reconstruction module $\mathsf{M}^r$ that uses image-caption embeddings $(\mathbf{I}, \mathbf{C})$ to reconstruct the embedding of the original truthful caption $\hat{\mathbf{C}}^t$. 
The reconstructed embedding serves as an additional signal for the detection model: alignment between $\hat{\mathbf{C}}^t$ and the input $\mathbf{C}$ indicates a likely truthful pair, 
while significant divergence may suggest manipulation.
We explore two training strategies—end-to-end and large-scale pre-training—and investigate mechanisms like attention, gating, and masking for integration. As shown in Fig.~\ref{fig:lamar}, including a backbone encoder, modality fusion, reconstruction, and detection modules.

\subsection{Backbone Encoder}

We use CLIP ViT L/14 from OpenCLIP \footnote{\url{https://github.com/mlfoundations/open_clip}} as the backbone multimodal encoder, $\mathsf{E}_I$ and $\mathsf{E}_C$, to produce image embeddings $\mathbf{I} \in \mathbb{R}^{768\times 1}$ and text embeddings $\mathbf{C} \in \mathbb{R}^{768\times 1}$, which are pre-aligned within a shared embedding space. 

\subsection{Modality Fusion}

To effectively integrate visual and textual modalities, we employ a fusion strategy that combines concatenation ($;$) with element-wise vector operations (addition, subtraction, and multiplication). 
This approach has been shown to be a lightweight yet effective method for capturing complementary relationships and differences between the two modalities \cite{papadopoulos2023red}. 
Specifically, we define the fused representation $\mathbf{F}$:
\begin{equation}
\label{eqn:fusion}
\mathbf{F} = [\mathbf{I};\mathbf{I}+\mathbf{C};\mathbf{I}-\mathbf{C};\mathbf{I}*\mathbf{C};\mathbf{C}]
\end{equation}
with $\mathbf{F}\in{R}^{5\times768}$. 
Concatenation $[\mathbf{I}; \mathbf{C}]$ provides the original, isolated features, while the element-wise operations $\mathbf{I}+\mathbf{C}$, $\mathbf{I}-\mathbf{C}$, and $\mathbf{I}*\mathbf{C}$ are included to highlight complementarity, differences, and mutual agreement between the visual and textual embeddings, respectively.

\subsection{Detection Network}

For the detection model $\mathsf{M}^d$, we define a neural network as follows:
\begin{equation} 
\label{eqn:detection} 
\hat{y} = \textbf{W}_1 \cdot \text{GELU}(\textbf{W}_0 \cdot \text{Flatten}([\mathbf{F}; \mathbf{\hat{C}^t}])) 
\end{equation}

where $\textbf{W}_0 \in \mathbb{R}^{768 \times 768}$ is a fully connected layer followed by a GELU activation, and $\textbf{W}_1 \in \mathbb{R}^{n \times 768}$ is the final classification layer, with $n=1$ for binary classification or $n=3$ multi-class classification. 
Bias terms $b$ are included in the model but omitted here for brevity.
Here, the operation `Flatten' refers to converting the concatenated vector $[\mathbf{F}; \mathbf{\hat{C}^t}]$ into a one-dimensional vector before passing it through the hidden layers.
The Detection Network is optimized using binary cross-entropy loss for binary classification and categorical cross-entropy for multi-class classification; denoted as $\mathcal{L}_d$.

\subsection{Reconstruction Network}

For the reconstruction network $\mathsf{M}^r$, we follow prior research in using Transformer encoder $\mathsf{T(\cdot)}$ for MMD \cite{papadopoulos2023synthetic, papadopoulos2024verite, papadopoulos2023red, papadopoulos2024similarity}, formulated as:
\begin{equation}
\label{eqn:transformer}
[\mathbf{t}_{\text{CLS}}, \mathbf{t}_{F}] = \mathsf{T}([\mathbf{CLS}; \mathbf{F}]), \quad \mathbf{\hat{C}^t} = \mathbf{t}_{\text{CLS}}
\end{equation}
where $CLS$ is a trainable classification token that serves as a global representation of all inputs, and its transformation is defined as the reconstructed caption embedding $\mathbf{\hat{C}^t}$.
The network is optimized using the MSE loss function, as defined in Eq. \ref{eqn:loss}.
Both input and target embeddings are L2-normalized, constraining representations to a unit hypersphere, allowing the MSE loss to focus on semantic alignment rather than magnitude variations.

\subsubsection{End-to-end (E2E) training} 
To jointly optimize the reconstruction network $\mathsf{M}^r$ and the detection model $\mathsf{M}^d$, we explore end-to-end multi-task training, where the entire LAMAR model is trained using both reconstruction loss $\mathcal{L}_{r}$ (e.g., MSE) and detection loss $\mathcal{L}_{d}$ (binary or categorical cross-entropy) as $\mathcal{L} = \mathcal{L}_{d} + \mathcal{L}_{r}$. 

Under this setting, the input caption embedding $\mathbf{C}$ may correspond to either a truthful or a manipulated caption, requiring $\mathsf{M}^r$ to learn how to handle both cases appropriately.
Since $\mathsf{M}^r$ and $\mathsf{M}^d$ are trained jointly, the detection model can more dynamically modulate the influence of the reconstructed embedding based on its semantic alignment with the inputs.
Specifically, we investigate:

\noindent
\textit{1.1)}
Direct integration of the reconstructed caption embedding $\mathbf{\hat{C}^t}$ into $\mathsf{M}^d$, as shown in Eq.\ref{eqn:detection}.

\noindent
\textit{1.2)}     
A gating mechanism, formulated as: 
\begin{equation}
\label{eqn:gate}
\mathbf{g} = \mathcal{S}(\mathbf{W}_g \cdot \mathbf{F} + \mathbf{b}_g)
\end{equation}
\begin{equation}
\label{eqn:gate_2}
\mathbf{\hat{C}^t}_{gate} = \mathbf{g} \odot \mathbf{\hat{C}^t}
\end{equation}
where $\mathbf{g}$ is the gate, $S$ the sigmoid function and $\mathbf{W_g} \in \mathbb{R}^{768 \times 768 }$ while the input to Eq.\ref{eqn:detection} is altered to $[\mathbf{F}; \mathbf{\hat{C}^t}_{gate}]$.

\noindent
\textit{1.3)} 
A masking mechanism, formulated as:
\begin{equation}
\label{eqn:mask}
\mathbf{m} \sim Bernoulli(\mathcal{S}(\mathbf{W}_m \cdot \mathbf{F} + \mathbf{b}_m))
\end{equation}
\begin{equation}
\label{eqn:mask_2}
\mathbf{\hat{C}^t}_{mask} = \mathbf{m} \odot \mathbf{\hat{C}^t}
\end{equation}
where $\mathbf{m}$ represents binary mask sampled from a Bernoulli distribution, $\mathbf{W_m} \in \mathbb{R}^{768 \times 768}$
and the input to Eq.\ref{eqn:detection} is altered to $[\mathbf{F}; \mathbf{\hat{C}^t}_{mask}]$.

\noindent
\textit{1.4)} 
An attention mechanism, formulated as: 
\begin{equation}
\label{eqn:attention}
\mathbf{F_a} = [\mathbf{I}, \mathbf{C}, \mathbf{\hat{C}}]
\end{equation}
\begin{equation}
\label{eqn:attention_2}
\mathbf{Q} = \mathsf{W}_Q \cdot \mathbf{F_a}, 
\quad \mathbf{K} = \mathsf{W}_K \cdot \mathbf{F_a}, 
\quad \mathbf{V} = \mathsf{W}_V \cdot \mathbf{F_a}
\end{equation}
\begin{equation}
\label{eqn:attention_3}
\mathbf{\hat{C}^t}_{attend} = mean(softmax\left(\frac{\mathbf{Q} \cdot \mathbf{K}^T}{\sqrt{768}}\right) \cdot \mathbf{V})
\end{equation}
where 
$\mathsf{W}_Q, \mathsf{W}_K, \mathsf{W}_V \in \mathbb{R}^{768 \times 768}$, and $mean$ denotes average pooling across the first dimension.  
The input to Eq.\ref{eqn:detection} is altered to $[\mathbf{F}; \mathbf{\hat{C}^t}_{attend}]$.

\subsubsection{Large-scale Pre-Training (PT)}
In addition to end-to-end training, we also investigate a large-scale pre-training approach where the reconstruction network is trained exclusively on truthful captions using a large-scale image-caption dataset, VisualNews \cite{liu2021visual}, comprising 1,259,732 truthful image-caption pairs. 
Unlike end-to-end training, in this setting $\mathsf{M}^r$ is trained separately using only perturbed captions $C^f$ as input. We explore two pre-training strategies:

\noindent
\textit{2.1)} Gaussian noise is added to the original text embedding, and the network is tasked with reconstructing the noisy embedding; expressed as: 
\begin{equation}
\label{eqn:gauss} 
\mathbf{C}^f = \mathbf{C}^t + \mathcal{N}(\mu, \sigma^2) 
\end{equation}
where $\mathcal{N}(\mu, \sigma^2)$ denotes Gaussian noise with mean $\mu$ and standard deviation $\sigma$.

\noindent
\textit{2.2)} 
Similarly, dropout is applied to the original text embedding, and the network is tasked with reconstructing the dropped-out embedding; expressed as: 
\begin{equation} 
\label{eqn:dropout} 
\mathbf{C}^f = Dropout(\mathbf{C}^t, dp) \end{equation} where $dp$ denotes the dropout probability. 

In both cases, the noisy or dropout-modified embedding $\mathbf{C^f}$ is then substituted for $\mathbf{C}$ in Eq.~\ref{eqn:fusion} and used in $\mathsf{M}^r$ to reconstruct the truthful caption embedding $\mathbf{\hat{C}}^t$. 
Once $\mathsf{M}^r$ is trained, the reconstructed embeddings are integrated into the detection network $\mathsf{M}^d$ during its training. 
To integrate these embeddings, we explore direct integration as well as the gate and attention mechanisms.

\section{Experimental Setup}
\label{sec:setup}

\subsection{Training Datasets}
\label{sec:datasets}

For the ``True vs. MC'' experiments, we use the 4 versions produced with LLaVA, 1 from Llama, 1 from Molmo, as detailed in Section~\ref{sec:dataset}, along with two additional datasets: the deduplicated version of the Crossmodal HArd Synthetic MisAlignment (CHASMA) dataset, which includes 145,891 truthful and 145,891 miscaptioned images \cite{papadopoulos2024verite}, and the CLIP-based Named Entity Swapping by Topic (NESt) dataset, containing 847,693 miscaptioned images and 1,007,744 truthful images \cite{papadopoulos2023synthetic}.
For ``True vs. OOC'' experiments, we use the NewsCLIPpings dataset \cite{luo2021newsclippings}, Merged/Balanced version, comprising 42,680 truthful and OOC samples in total. 
Finally, for multi-class classification, we combine one of the ``miscaptioned images'' datasets (\datasetname, CHASMA, or NESt) with the NewsCLIPpings dataset, which represents the OOC class, and apply random under-sampling to balance the classes.

\subsection{Evaluation Protocol}

We adhere to the training, validation, and testing splits provided by each dataset and evaluate models using Accuracy as the primary metric.
After training on any of the aforementioned synthetic datasets, we further assess performance on the VERITE benchmark \cite{papadopoulos2024verite}, which comprises 1,000 real-world samples: 338 truthful image-caption pairs, 338 miscaptioned images, and 324 out-of-context pairs.
We report `True vs. OOC' and `True vs. MC' accuracies for binary classification tasks, and overall accuracy for the multi-class task.
All three tasks serve as out-of-distribution evaluations, as the training data are synthetically generated, while the final evaluation is conducted on real-world data.

For the `True vs. MC' task, we update the VERITE benchmark, following the data selection criteria and labeling process defined in $\cite{papadopoulos2024verite}$. 
Specifically, we collected 50 new samples -- sourced from Snopes and Reuters, balanced across the two classes -- describing events that occurred between October 2024 and July 2025.
Crucially, these data points fall after the knowledge cut-off point of the selected VLMs; Molmo, Llama, and LLaVA. 
This expansion, which we denote \textit{VERITE 24/25}, was collected to examine the out-of-distribution robustness and temporal generalization capacity of both zero-shot VLMs and our proposed methods against emerging misinformation.

\subsection{Competing Methods}

In addition to our proposed LAMAR architecture and its variations, we reproduce and evaluate the performance of several competing methods:
(1) \textbf{MUSE} \cite{papadopoulos2024similarity}: a simple similarity-based baseline consisting of a single hidden layer applied to cosine similarity scores between image-text pairs. 
(2) \textbf{SpotFake*} $\cite{singhal2019spotfake}$: A neural network that processes the input sequence $[\mathbf{I}; \mathbf{C}]$.
(3) SAFE* (Similarity-Aware FakE) $\cite{zhou2020similarity}$: Similar to \textbf{SpotFake*}, but processing the input sequence $[\mathbf{I}; \mathbf{C}; sim(\mathbf{I}, \mathbf{C})]$.
(4) \textbf{MVAE*} (Multimodal Variational Autoencoder) \cite{khattar2019mvae}: A bimodal variational autoencoder that maps image and text features into a shared latent space. The network is trained to jointly minimize the reconstruction, KL divergence, and cross-entropy losses. 
The decoder reconstructs the original input words and the image embeddings from the shared latent space. 
(5) \textbf{BCMF*} (Bidirectional Cross-Modal Fusion) \cite{yu2022bcmf}: A dual-stream architecture that performs bidirectional cross-attention between images and captions, followed by gating mechanisms to refine the features. 
The final fused feature vector concatenates the refined features and their explicit interaction terms, expressed as $[\text{Maxpool}(\mathbf{C}') ; \mathbf{I}' ; |\text{Maxpool}(\mathbf{C}') - \mathbf{I}'|]$, which is then used for the final classification.
(6) \textbf{PMFN*} (Progressive MLP-Mixer Fusion Network) \cite{jing2023multimodal}: A multi-stage architecture, that performs hierarchical fusion. 
It leverages the MLP-Mixer module to iteratively refine multimodal features by fusing image and text representations from different abstraction stages of the backbones, aiming to enhance the correlation between shallow and deep features before final classification. 
(7) \textbf{DT-Transformer} \cite{papadopoulos2023synthetic}: a Transformer encoder that processes the input sequence $[\mathbf{CLS}; \mathbf{I}; \mathbf{C}]$. 
(8) \textbf{RED-DOT} \cite{papadopoulos2023red}: a Transformer-based model operating on the fused representation $[\mathbf{CLS}; \mathbf{F}]$; we use the Baseline version without external evidence. 
(9) \textbf{AITR} (Attentive Intermediate Transformer Representations) \cite{papadopoulos2024similarity}: A model that uses a stack of $n$ Transformer encoder layers of varying number of attention heads to process the initial multimodal sequence $[\mathbf{CLS}_0; \mathbf{F}]$ and then applies a self-attention mechanism over the sequence of intermediate classification tokens $[\mathbf{CLS}_1, \dots, \mathbf{CLS}_n]$ for final classification.

For a fair comparison, all models marked with an asterisk (*) were adapted to use CLIP ViT-L/14 image and text features, replacing their original backbones (VGG-19, BERT, DeiT, Swin Transformer, Bi-LSTM, etc.).

\subsection{Implementation Details}

We train LAMAR using a Transformer encoder with 4 encoder blocks, each containing 4 multi-head self-attention heads and a feed-forward layer of 1,024 dimensions and a dropout of 0.1 probability.  
The model is optimized with Adam, using a learning rate of $1e-4$ and a batch size of 512. 
We train the network for up to 50 epochs, with early stopping after 10 epochs if validation performance does not improve. 
For the large-scale pre-training, we consider $\sigma \in \{0.1, 0.2 \}$ and $\mu = 0.0$ and dropout probability $dp \in \{ 0.2, 0.5\}$. 
To ensure reproducibility, we set a constant random seed (0) for PyTorch, Python Random, and NumPy.

\subsection{Computational Complexity}

Generating a \datasetname\ dataset with LLaVA, took 18.3 hours using a single Nvidia GeForce RTX 4090 (24GB RAM). 
Feature extraction with CLIP ViT-L/14 required approximately 16 minutes for images and 2 minutes for texts, using a batch size of 256; and was performance only once per dataset, and is used for any model. 

Using fvcore\footnote{\url{https://github.com/facebookresearch/fvcore}}, 
we estimate that LAMAR, trained end-to-end with a gate mechanism, requires 980,997 FLOPs, and 1,034,181 FLOPs with the attention mechanism. 
This is comparable to AITR (1,070,653 FLOPs) \cite{papadopoulos2024similarity} and more efficient than RED-DOT (2,208,072 FLOPs) \cite{papadopoulos2023red}.
When using pre-extracted features, LAMAR requires a maximum of 30 seconds per epoch (batch size 512) to process the unfiltered ($l=None$) dataset for multiclass classification.

\section{Results}
\label{sec:results}

\begin{figure*}[!t]
\centering
\includegraphics[width=\textwidth]{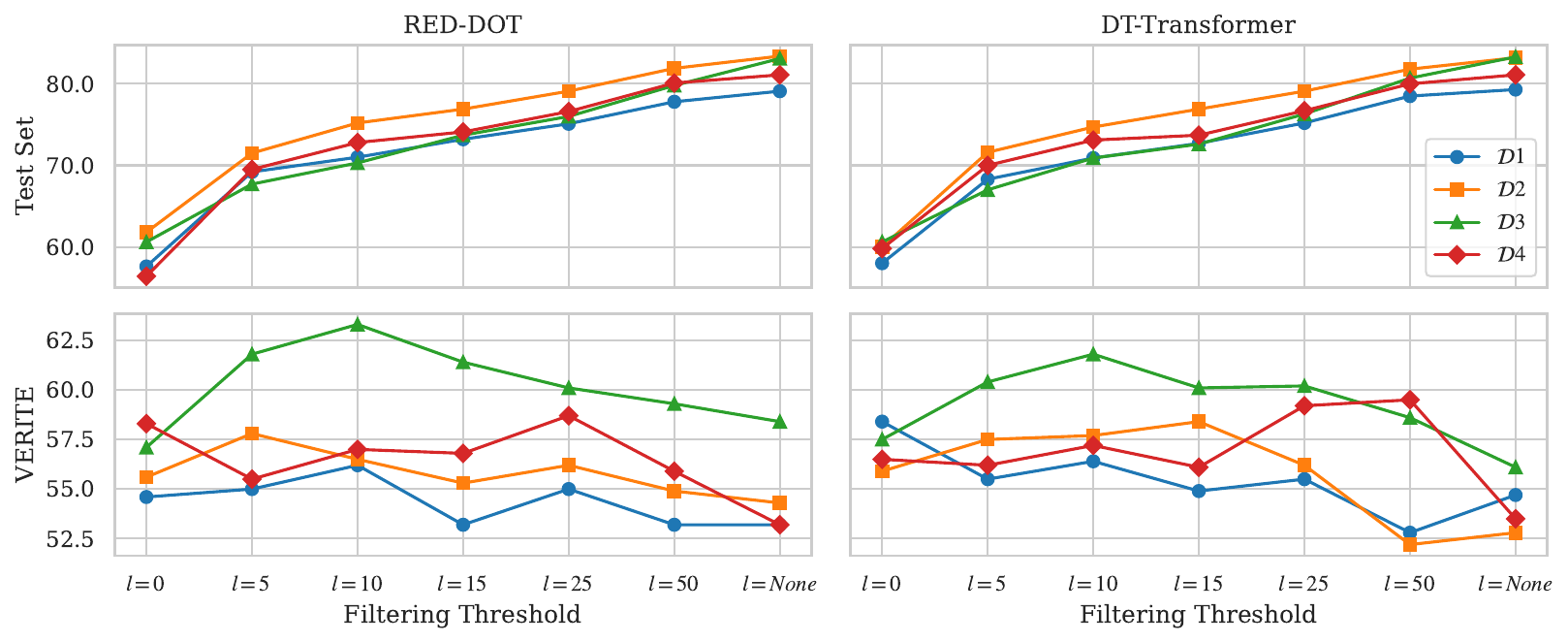}
\caption{
Performance of detection models (DT-Transformer and RED-DOT) trained on four variations of LLaVA-$\mathcal{D}_1$, $\mathcal{D}_2$, $\mathcal{D}_3$, and $\mathcal{D}_4$, 
evaluated under varying filtering thresholds ($l \in \{0,5,10,15,25,50,\text{None}\}$, or 4.5\%, 19.1\%, 27.8\%, 34.9\%, 47.9\%, 74.0\%, and 100\% of the dataset) in terms of Test-set Accuracy and VERITE ``True vs. MC'' Accuracy.
}
\label{fig:filter}
\end{figure*}

\subsection{Dataset Variants and Filtering}

First, we investigate how different synthetic data variations ($\mathcal{D}_1$--$\mathcal{D}_4$) and filtering thresholds ($l$) influence a model's ability to detect synthetic and real-world misinformation. 
Specifically, we analyze the relationship between internal (synthetic) test-set performance and out-of-distribution generalization to the VERITE (``True vs. MC'') benchmark.
As shown in Fig.~\ref{fig:filter}, test-set accuracy tends to improve with higher $l$ values across all datasets and two established MMD models, DT-Transformer \cite{papadopoulos2023synthetic} and RED-DOT \cite{papadopoulos2023red}.
The highest test-set performance is observed without any filtering ($l = \text{None}$), ranging from 79\% to 83.4\%. 
However, we identify a divergence between test-set and real-world performance: with a negative Pearson correlation of $-0.29$ for DT-Transformer and $-0.17$ for RED-DOT.
This indicates that high test-set accuracy on synthetic data does not guarantee better generalization; rather, training on synthetic misinformation that is too easily distinguishable from truthful pairs likely encourages the model to exploit spurious shortcuts rather than learning robust factual consistency, ultimately hindering its generalization.

Regarding filtering, both models achieve peak performance at $l=10$, with accuracies of 63.3\% and 61.8\%, followed by $l=5$, $15$, and $25$. 
This highlights the importance of filtering out the ``rambling'' or generic descriptions occasionally produced by VLMs. 
Moreover, filtering overly long texts prevents the detection model from relying on superficial length-based patterns (e.g., that longer texts tend to be ``falsified'') rather than actual factual consistency.
Thus, by refining the dataset in this way, we improve model robustness and enhance real-world generalization on VERITE.

Overall, both models achieve their highest performance when trained on $\mathcal{D}_3$, with RED-DOT and DT-Transformer averaging 60.2\% and 59.2\%, respectively. 
This suggests that $p_3^{gen}$ produces misinformation patterns that more closely resemble real-world examples, resulting in better model generalization.
Consequently, we adopt $p_3^{gen}$ to produce \textit{Molmo-}$\mathcal{D}_3$ and \textit{Llama-}$\mathcal{D}_3$, and use \textit{LLaVA-}$\mathcal{D}_3$ variant of \datasetname\ for proceeding experiments, while treating $l$ as a hyperparameter.

\subsection{Ablation across Datasets}

\begin{table}[t!]
    \centering
    \caption{LAMAR ablations trained on three synthetic datasets (LLaVA-$\mathcal{D}_3$, NESt, or CHASMA) and evaluated on VERITE (``True vs. MC''). 
    \textbf{Bold} denotes the highest accuracy.
    }    
    \renewcommand{\arraystretch}{1.2} 
    \begin{tabular}{lccc}  
        \hline

        \hline
        \textbf{LAMAR Variant} & \textbf{LLaVA-$\mathcal{D}_3$} & \textbf{NESt} & \textbf{CHASMA} \\
        
        \hline

        E2E, Gate & \textbf{66.0 }& \textbf{61.2} & \textbf{59.8} \\
        E2E, Attention & 65.1 & 58.6 & 58.6\\
        E2E, Mask & 63.6 & 59.5 & 58.1\\

        E2E, Gate - Fusion & 61.1 & 57.3 & 58.0 \\
        E2E, Attention - Fusion & 61.2 & 55.0 & 57.0 \\

        E2E, Direct & 63.8  & 59.2 & 57.8\\
        E2E, Direct - Image  & 63.2 & 58.3 & 57.3\\        
        E2E, Direct - Text & 63.0 & 58.7 & 57.1\\       

        \hline 

        PT, Dropout, Gate & 63.0 & 57.5 & 58.6 \\
        PT, Dropout, Attention & 64.5 & 57.1 & 56.5\\
        PT, Dropout, Direct & 61.2 & 59.5 & 57.1\\       

        PT, Gaussian, Gate & 62.1 & 58.7 & 59.3\\
        PT, Gaussian, Attention & 63.0 & 59.3 & 57.5\\
        PT, Gaussian, Direct & 63.2 & 58.6 & 57.5 \\       

        \hline
        No Reconstruction & 62.7 & 58.2 & 56.7\\
        \hline
    \end{tabular}

    \label{tab:ablation}
\end{table}

We conduct an extensive analysis of the proposed LAMAR method, including various model variants and ablations. This analysis is conducted using three distinct synthetic datasets for training: \datasetname\ (LLaVA-$\mathcal{D}_3$), CHASMA \cite{papadopoulos2024verite}, and NESt \cite{papadopoulos2023synthetic}. The performance of all models is consistently evaluated on the VERITE benchmark, specifically, the ``True vs. MC'' task. 

As shown in Table \ref{tab:ablation}, training on \datasetname\ (LLaVA-$\mathcal{D}_3$) consistently yields models with stronger generalization to real-world data. This trend holds across all LAMAR variants and supports our hypothesis that VLM-generated training data provides more effective supervision than earlier methods based on named entity swaps or cross-modal inconsistencies.
The best overall performance (66\%) is achieved by LAMAR [E2E] with the Gate integration mechanism trained on \datasetname\ data (LLaVA-$\mathcal{D}_3$), outperforming NESt by 7.8\% and CHASMA by 10.4\%. 
E2E training proves advantageous, as it directly optimizes reconstruction over actual miscaptioned pairs, unlike PT variants that rely on perturbed truthful captions via noise or dropout.
These results highlight the value of semantically grounded, VLM-generated training data that more closely resembles real-world misinformation.

Among integration strategies, \textit{Gate} and \textit{Attention} mechanisms consistently outperform direct embedding injection, highlighting the importance of adaptive integration. 
In contrast, the Mask mechanism underperforms, likely due to reduced representational capacity from fully masking the reconstruction embedding. 
Similarly, removing the fusion component $\mathbf{F}$ (Eq.~\ref{eqn:fusion}) and using simple concatenation ($[\mathbf{I};\mathbf{C}]$) significantly degrades performance, emphasizing the role of effective multimodal fusion.

We also conduct unimodal ablations. Removing the image input (Direct-Image) leads to a slight drop in performance (-0.9\%), while removing the text (Direct-Text) results in a larger drop (-1.3\%), effectively turning the task into latent captioning. 
Finally, eliminating the reconstruction module entirely yields the lowest E2E performance, confirming the critical role of the reconstruction in guiding accurate detection.

To evaluate how LAMAR integrates the reconstructed signal with the primary inputs, we analyze the distribution of attention scores across input tokens (image, caption, and reconstructed caption embeddings) and the behavior of the gating mechanism.
Attention scores are substantially higher on the reconstruction token compared to the image or input caption tokens across classes. Truthful pairs receive higher attention on average for both the image token (0.218 vs. 0.176) and the input caption (0.213 vs. 0.169), while falsified pairs shift attention toward the reconstructed caption (0.655 vs. 0.569).
Similarly, the gate score correlates positively with reconstruction loss (0.451), and reconstruction loss itself is positively associated with the model's predicted label (0.386), suggesting that higher reconstruction error pushes the model toward a ``falsified'' prediction.
These patterns indicate that LAMAR learns to focus more on the reconstructed caption, via attention or gating, when detecting inconsistencies, making it a valuable auxiliary signal for accurate classification.
Thus, when the reconstruction caption embeddings diverges semantically from the input embeddings, the model flags the pair as falsified.

\subsection{Comparative Study across VLM-Generated Datasets}
Table~\ref{tab:comparative_mc} shows the performance of LAMAR and prior SotA models trained on VLM-generated datasets produced by LLaVA, Molmo, and Llama, and evaluated on the VERITE benchmark (``True vs. MC'').
Across all three datasets, LAMAR consistently outperforms prior models by significant margins: 14.4\% over MUSE, 8.7\% over MVAE, 7.0\% over SpotFake, 6.3\% over SAFE, 3.8\% over BCMF, 3.8\% over MPFN, 6.8\% over DT-Transformer, 4.3\% over RED-DOT, and 4.9\% over AITR.
This holds for both LAMAR variants using gated or attention-based integration.
The best overall result is achieved by LAMAR [E2E, Gate] trained on \textit{LLaVA-}$\mathcal{D}_3$, reaching 66.0\% accuracy.

\begin{table}[t!]
    \centering
    \caption{Performance of models trained on three VLM-generated datasets (\textit{LLaVA}-$\mathcal{D}_3$, \textit{Molmo-}$\mathcal{D}_3$, \textit{Llama-}$\mathcal{D}_3$) and evaluated on VERITE (``True vs. MC'').
\textbf{Bold} indicates the highest overall accuracy; \underline{Underline} indicates the best accuracy per training dataset.
    }    
    \renewcommand{\arraystretch}{1.2}
    \begin{tabular}{lccc}  
        \hline

        \textbf{Model} & \textbf{LLaVA-$\mathcal{D}_3$} & \textbf{Molmo-}$\mathcal{D}_3$ & \textbf{Llama-}$\mathcal{D}_3$ \\
        \hline
        MUSE \cite{papadopoulos2024similarity} & 57.7 & 54.9 & 57.5 \\

        MVAE \cite{khattar2019mvae} & 60.7 & 60.8 & 60.5 \\

        SpotFake \cite{singhal2019spotfake} & 61.7 & 62.6 & 60.7 \\
        SAFE \cite{zhou2020similarity} & 62.1 & 63.0 & 59.6\\      

        BCMF \cite{yu2022bcmf} & 63.6 & 61.2 & 59.0 \\

        MPFN \cite{jing2023multimodal} & 63.6 & 64.1 & 58.6 \\
        
        DT-Transformer \cite{papadopoulos2023synthetic} & 61.8 & 62.1 & 58.4\\
        RED-DOT \cite{papadopoulos2023red} & 63.3 & 62.9 & 61.7\\
        AITR \cite{papadopoulos2024similarity} & 62.9 & 63.2 & 63.0\\        

        \hline

        LAMAR [E2E, Attention] & 65.1 & \underline{65.1} & \underline{64.2}\\          
        LAMAR [E2E, Gate] & \textbf{\underline{66.0}} & 64.4 &  63.9\\          

        \hline
    \end{tabular}

    \label{tab:comparative_mc}
\end{table}

LLaVA-generated data yielding higher performance may be partially attributed to the use of ``adversarial prompt selection'' on it, whereas Molmo and Llama data were generated using the same prompt ($p_3^{gen}$). 
While this introduces a slight advantage for LLaVA, it is notable that models trained on Molmo- and Llama-generated data performed comparably well, thus underscoring the general utility of VLM-generated data across models.

Qualitative analysis (see Fig.~\ref{fig:examples}) further highlights that LLaVA-$\mathcal{D}_3$ contain more nuanced and believable false captions by maintaining semantic consistency with the image. 
In contrast, LLaVA-$\mathcal{D}_4$ and Llama-$\mathcal{D}_3$ often contain more apparently false captions -- for example, attributing protests to ``Madrid, Spain'' or ``in front of the Eiffel Tower in Paris'' despite clear visual cues such as Greek flags. 
Such easily detectable falsehoods are less reflective of real-world misinformation and thus less effective for training models that generalize well.

\subsection{Out-of-context and Multiclass Detection}

\begin{table}[t!]
    \centering
    \caption{Performance of models on the ``True vs. OOC'' task, trained on the NewsCLIPpings dataset and evaluated on both NewsCLIPpings and VERITE.}
    \renewcommand{\arraystretch}{1.2}
    \begin{tabular}{lcc}  
        \hline
        \textbf{Model} & NewsCLIPpings & VERITE \\
        \hline
        CLIP \cite{luo2021newsclippings} & 60.2 & - \\
        SSDL \cite{mu2023self} & 71.0 & -\\

        MVAE \cite{khattar2019mvae} & 60.6 & 55.2 \\
        
        MUSE \cite{papadopoulos2024similarity} & 80.6 & 72.0\\        
        SpotFake \cite{luo2021newsclippings} & 79.8 & 67.1 \\        
        SAFE \cite{zhou2020similarity} & 83.7 & 72.7 \\      

        BCMF \cite{yu2022bcmf} & 76.6 & 67.1 \\

        MPFN \cite{jing2023multimodal} & 76.1 & 66.5\\

        DT-Transformer \cite{papadopoulos2023synthetic} & 79.7 & 69.4 \\        
        RED-DOT \cite{papadopoulos2023red} & 81.5 & 73.5 \\
        AITR \cite{papadopoulos2024similarity} & 84.1 & 74.1 \\
        
        \hline

        LAMAR [E2E, Attention] & \textbf{84.8} & 75.1  \\
        LAMAR [E2E, Gate] & 84.7 & \textbf{76.3} \\
        
        \hline
    \end{tabular}

    \label{tab:ooc}
\end{table}

\begin{table}[t!]
    \centering
    \caption{Comparison of multi-class models trained either on CHASMA, \textit{LLaVA}-$\mathcal{D}_3$, \textit{Molmo-}$\mathcal{D}_3$, and evaluated on VERITE.
    }
    \renewcommand{\arraystretch}{1.2} 
    \begin{tabular}{lccc}
        \hline
        \textbf{Model} & \textbf{CHASMA} &
        \textbf{\textit{LLaVA}}-$\mathcal{D}_3$ & \textbf{\textit{Molmo}-$\mathcal{D}_3$} 
        \\
        \hline
        MUSE \cite{papadopoulos2024similarity} & 38.5 & 49.7 & 47.3\\

        MVAE \cite{khattar2019mvae} & 39.8 & 40.5 & 42.7 \\
        
        SpotFake \cite{singhal2019spotfake} & 49.2 & 48.7 & 47.2  \\

        SAFE \cite{zhou2020similarity} & 50.4 & 51.4 & 51.0  \\

        BCMF \cite{yu2022bcmf} & 45.9 & 45.4 & 47.5 \\

        MPFN \cite{jing2023multimodal} & 46.1 & 44.1 & 44.8\\

        DT-Transformer \cite{papadopoulos2023synthetic} & 50.0 & 47.8 & 48.7  \\
        RED-DOT \cite{papadopoulos2023red} & 48.5 & 48.8 & 49.1 \\
        AITR \cite{papadopoulos2024similarity} & 51.4 & 51.7 & 52.0  \\
        LAMAR [E2E, Gate] & 53.2 & 54.2 & \textbf{54.9} \\
        \hline
    \end{tabular}
    \label{tab:multiclass}
\end{table}

Table \ref{tab:ooc} presents LAMAR's performance compared to prior SotA models on ``True vs. OOC'' classification, without access to external evidence. 
LAMAR achieves the highest accuracy on the NewsCLIPpings dataset, with [E2E, Attention] reaching 84.8\% and [E2E, Gate] closely following at 84.7\%, both outperforming AITR (84.1\%) and other models.
More importantly, LAMAR [E2E, Gate] achieves 76.3\% on VERITE, surpassing the best prior model (AITR) by 3\%.
We observe a consistent performance gap between True vs. OOC'' (76.3\%) and True vs. MC'' (66.0\%), reflecting the greater difficulty of detecting subtle manipulations (e.g., dates, places, or actions) in MC cases, compared to the more blatant mismatches typical of OOC examples.

For the multi-class classification task, 
we employ three training datasets: \textit{LLaVA-}$\mathcal{D}_3$, \textit{Molmo-}$\mathcal{D}_3$, and \textit{CHASMA}; all being combined with NewsCLIPpings to represent the OOC class.
As shown in Table~\ref{tab:multiclass}, 
LAMAR [E2E, Gate] consistently generalizes best across all training sets.
When trained on \textit{Molmo-}$\mathcal{D}_3$, LAMAR achieves the highest overall accuracy (54.9\%), representing a relative improvement of 5.6\% over AITR.
These results further validate the benefits of using VLM-generated data in LAMAR.  

\begin{table}[t!]
    \centering
    \caption{Comparison of zero-shot VLMs against fine-tuned LAMAR variants on the ``True vs. MC'' task of \textit{VERITE} and the post-knowledge-cutoff data of \textit{VERITE 24/25}.
    }
    \begin{tabular}{lccc}
        \hline
        \textbf{Model} & \textbf{Parameters} & \textbf{VERITE} & \textbf{VERITE 24/25} \\
        \hline
        Llama 3.2 & 11B & 59.3 & 53.1 \\
        Molmo & 7B & 63.0 & 58.0 \\
        LLaVA 1.6 & 7B & \textbf{66.9} & 62.0 \\
        
        \hline
        
        LAMAR [E2E, Attention] & 21M & 65.1 & \textbf{70.0} \\
        LAMAR [E2E, Gate] & 19M & 66.0 & 68.0 \\

        \hline
    \end{tabular}
    \label{tab:vlm_detection}
\end{table}

\subsection{Temporal Generalization Analysis}

Table \ref{tab:vlm_detection} compares the performance of the three VLMs used for the generation of \datasetname\ data, in a zero-shot setting against our fine-tuned LAMAR variants on the VERITE ``True vs. MC'' task. 
On the original VERITE dataset, the performance of the VLMs varies, with LLaVA 1.6 (66.9\%) showing comparable performance to LAMAR [E2E Gate] (66.0\%) and outperforming Molmo (63.0\%) and Llama 3.2 (59.3\%). 
Decent zero-shot performance is not unexpected, as these VLMs are trained on vast, large-scale datasets, making it highly probable that they have encountered samples or related content from popular fact-checking sources like Snopes and Reuters used in the original VERITE benchmark. 

However, the analysis shifts when evaluated on the VERITE 24/25 set comprising samples dating after the VLMs' knowledge cut-off. 
Here, all zero-shot VLMs demonstrate a notable decline in performance, with LLaVA 1.6 dropping to 62.0\%, Llama 3.2 falling to 53.1\%, and Molmo to 58.0\%. 
In contrast, both fine-tuned LAMAR variants demonstrate strong temporal generalization, with LAMAR [E2E Attention] achieving 70.0\% and LAMAR [E2E Gated] 68.0\% on the post-cutoff data, despite having only $\approx$20 million parameters compared to the 7-11 billion parameters of the VLMs.
The performance gap on VERITE 24/25 highlights that VLM's zero-shot performance is limited by their training data and knowledge cutoff, whereas LAMAR's reconstruction-based approach provided higher robustness against emerging misinformation.

\section{Ethical Discussion}

Our work demonstrates the potential of VLMs for generating high quality training datasets for multimodal misinformation detection.
Nevertheless, it is crucial to consider the ethical implications of this work and future research in this direction. 
In our initial attempts, earlier open-source VLMs such as MiniGPT and LLaVA 1.5 struggled to generate coherent or realistic false captions. 
However, later versions (LLaVA 1.6, Llama 3.2, and Molmo) were more capable of producing false but plausible captions. 
While this capability proved valuable for research -- enabling the creation of more robust training datasets -- it also raises serious societal concerns as even lightweight, open-source models can now be repurposed to automate and scale disinformation campaigns. 
We believe that the research community must remain vigilant, ensuring that advancements in detection do not inadvertently provide a blueprint for malicious generation. 
To mitigate these risks, we refrain from publicly releasing specific prompts used to generate the \datasetname\ data. 
However, to ensure scientific reproducibility, we provide the general templates within this work, and the exact prompts will be made available to verified researchers upon reasonable request for non-commercial research purposes.

\section{Conclusion}
\label{sec:conclusions}

In this study, we address the challenge of Multimodal Misinformation Detection (MMD) by introducing the \datasetname\ framework and the Latent Multimodal Reconstruction (LAMAR) architecture. 
\datasetname\ leverages Vision-Language Models (VLMs) to generate diverse, realistic synthetic miscaptioned images, significantly improving training data quality and ultimately model generalization. 
The LAMAR architecture adopts a reconstruction-based approach: it learns to reconstruct the embedding of the original, truthful caption from manipulated image-caption pairs, thereby offering an effective auxiliary signal that enhances detection. 
Extensive experiments demonstrate that models trained on \datasetname\ data significantly outperform previous methods -- including those based on named entity manipulation and cross-modal misalignment -- by 7.8\% and 10.4\% on the VERITE benchmark, respectively. 
Furthermore, LAMAR establishes new SotA performance across all three VERITE tasks: True vs. OOC (+4.3\%), True vs. MC (+3.0\%), and multiclass classification (+5.6\%).

Despite the promising gains achieved by \datasetname\ and LAMAR, this work stresses that MMD remains an open and challenging task. 
Apart from high ``out-of-context'' performance ($\approx 85\%$), our final model achieved $\approx 66\%$ and $\approx 55\%$ in the miscaptioned and multiclass scenarios, respectively, which indicates considerable room for improvement.
We argue that while assessing the internal consistency of the image-caption pair is a crucial prerequisite for MMD, it is rarely sufficient. 
Therefore, a critical future direction involves collecting and integrating external knowledge and evidence (e.g., related news articles) for comprehensive fact-checking. 
Although evidence-enriched datasets exist, such as the enriched NewsCLIPpings dataset for out-of-context cases \cite{abdelnabi2022open}, they often suffer from significant issues including information leakage, dataset-specific biases \cite{papadopoulos2024similarity}, or unreliable evidence $\cite{glockner2022missing, chrysidis2024credible}$. 
This emphasizes the pressing need for a new, robust, evidence-enriched dataset specifically targeting miscaptioned images to accurately model and address external inconsistency.
Synthetic data generation through \datasetname, and reconstruction-based architectures, such as LAMAR, could play an instrumental role in advancing these research directions.

\section*{Acknowledgments}
This work is partially funded by the projects ``vera.ai: VERification Assisted by Artificial Intelligence'' under grant agreement no. 101070093, ``DisAI - Improving scientific excellence and creativity in combating disinformation with artificial intelligence and language technologies'' under grant agreement no. 101079164, and ``AI4TRUST - AI-based-technologies for trustworthy solutions against disinformation'' under grant agreement no. 101070190. 

\bibliographystyle{unsrt}  
\bibliography{references}

\end{document}